\def\year{2022}\relax
\documentclass[letterpaper]{article} % DO NOT CHANGE THIS
\usepackage{aaai22}  % DO NOT CHANGE THIS
\usepackage{times}  % DO NOT CHANGE THIS
\usepackage{helvet}  % DO NOT CHANGE THIS
\usepackage{courier}  % DO NOT CHANGE THIS
\usepackage[hyphens]{url}  % DO NOT CHANGE THIS
\usepackage{graphicx} % DO NOT CHANGE THIS
\usepackage{natbib}  % DO NOT CHANGE THIS AND DO NOT ADD ANY OPTIONS TO IT
\usepackage{caption} % DO NOT CHANGE THIS AND DO NOT ADD ANY OPTIONS TO IT
\DeclareCaptionStyle{ruled}{labelfont=normalfont,labelsep=colon,strut=off} % DO NOT CHANGE THIS
\usepackage{algorithm}
\usepackage{algorithmic}

\usepackage[flushleft]{threeparttable}
\usepackage{array,booktabs,makecell,amsmath,amssymb,amsfonts,cite}
\usepackage{gensymb}
\usepackage{textcomp}
\usepackage{dblfloatfix}

\usepackage{newfloat}
\usepackage{listings}
\floatstyle{ruled}
\newfloat{listing}{tb}{lst}{}
\floatname{listing}{Listing}
\title{Computational Approaches for Modeling Power Consumption on an\\
Underwater Flapping Fin Propulsion System}
\author{
    Brian Zhou\textsuperscript{\rm 1} \thanks{This work was performed at the U.S. Naval Research Laboratory through a base program funded effort as part of the Science and Engineering Apprenticeship (SEAP) Program.},
    Jason Geder\textsuperscript{\rm 2},
    Alisha Sharma\textsuperscript{\rm 2,3},
    Julian Lee\textsuperscript{\rm 4},
    Marius Pruessner\textsuperscript{\rm 2},
    Ravi Ramamurti\textsuperscript{\rm 2},
    Kamal Viswanath\textsuperscript{\rm 2}
}
\affiliations {
    \textsuperscript{\rm 1} Thomas Jefferson High School, Alexandria, VA\\
    \textsuperscript{\rm 2} Naval Research Laboratory, Washington, DC\\
    \textsuperscript{\rm 3} University of Maryland College Park, College Park, MD\\
    \textsuperscript{\rm 4} Yale University, New Haven, CT\\
    2024bzhou@tjhsst.edu,
    \{jason.geder, alisha.sharma, marius.pruessner, ravi.ramamurti, kamal.viswanath\}@nrl.navy.mil,
    julian.lee@yale.edu
}

\begin{document}

\maketitle

\begin{abstract}
The last few decades have led to the rise of research focused on propulsion and control systems for bio-inspired unmanned underwater vehicles (UUVs), which provide more maneuverable alternatives to traditional UUVs in underwater missions. Propulsive efficiency is of utmost importance for flapping-fin UUVs in order to extend their range and endurance for essential operations. To optimize for different gait performance metrics, we develop a non-dimensional figure of merit (FOM), derived from measures of propulsive efficiency, that is able to evaluate different fin designs and kinematics, and allow for comparison with other bio-inspired platforms. We create and train computational models using experimental data, and use these models to predict thrust and power under different fin operating states, providing efficiency profiles. We then use the developed FOM to analyze optimal gaits and compare the performance between different fin materials. These comparisons provide a better understanding of how fin materials affect our thrust generation and propulsive efficiency, allowing us to inform control systems and weight for efficiency on an inverse gait-selector model. 
\end{abstract}

\section{Introduction}
Unmanned underwater vehicles (UUVs) have a variety of industrial and research applications including exploration, mapping, and minesweeping. Historically, these operations have primarily been conducted with propeller-driven UUVs. Because propeller-driven UUVs lack maneuverability and subsequently resistance to turbulence, their operational domain is limited to relatively quiescent and deeper waters. Marine animals offer promising solutions to expanding the envelope of UUV operations because they swim with high propulsive efficiency and have high maneuverability in water \cite{masud_estimate_2022, eloy_optimal_2012, taylor_flying_2003, rohr_strouhal_2004, rohr_observations_1998, triantafyllou_hydrodynamics_2000, nedelcu_underwater_2018}, which motivates replication of their fins and other appendages in robotic designs \cite{techet_propulsive_2008}. As such, recent decades have seen the rise of research in bio-inspired propulsion systems to fill the operational gap in littoral waters and create systems with greater agility and maneuverability compared to traditional propeller-based systems. Fin designs inspired by a variety of animals have been studied, including dolphins \cite{rohr_observations_1998, rohr_strouhal_2004}, penguins \cite{masud_estimate_2022}, and snakes. Among these, fish-inspired fins have driven the majority of research due to the agility these species exhibit, outperforming capabilities of traditional propulsion-based systems \cite{tangorra_development_2007, lauder_learning_2006, mignano_passing_2019, nedelcu_underwater_2018}.

Designing, optimizing, and replicating marine flapping fin motion with bio-inspired fins requires testing of different parameters. Previous studies have examined and tested the effects that different parameters such as material properties, kinematics or fin gaits, and fin shape have on a flapping fin's thrust output to better replicate and understand fish hydrodynamic performance \cite{sampath_hydrodynamics_2020, geder_maneuvering_2013, mignano_passing_2019, yun_thrust_2015, nguyen_thrust_2016, geder_underwater_nodate}. We identify and aim to resolve the following gaps in literature:
\begin{enumerate}
    \item Prior studies have placed emphasis on studying the effects of designs and gaits have on the thrust and lift forces \cite{sampath_hydrodynamics_2020, geder_maneuvering_2013, mignano_passing_2019, yun_thrust_2015, nguyen_thrust_2016, geder_underwater_nodate}. Less research has been conducted on the effects designs and flapping gaits have on power consumption and UUV efficiency. 
    \item Previous research on power has utilized CFD simulations to study hydrodynamic power \cite{palmisano_power_nodate}, but this is ineffective for control system integration that requires taking into account power loss of integrated actuators.
    %\item pectoral fin and low-speed propulsion
\end{enumerate}
Analyzing practical power draw and building models will allow us to create better design and gait recommendations while considering propulsive efficiency and integrating power onto a model-driven control system. This allows a control system to have different gait settings, such as a toggle-able setting optimizing for force output or gait efficiency, drastically increasing the possible mission duration for size-constrained littoral UUVs.

We explore the prediction of actuator power consumption and integration of these predictions into a control system for a set of fins on a UUV. Propeller-based systems have often been preferred due to both their repair-ability and efficiency \cite{palmisano_power_nodate}. While flapping fin systems may not exceed propellers in propulsive efficiency, it is essential for a prototype UUV to optimize for efficiency in thrust generation to ensure feasibility in a range of naval missions. Optimizing for efficiency is complex; previous research demonstrates that multiple variables affect thrust generation and efficiency in tandem fin configurations including flow speed and fin phase \cite{mignano_passing_2019}, as well as other stroke kinematics. Due to the popularity of bio-inspired fish fins, there have been many propulsors already created \cite{tangorra_development_2007}, but few have developed the models necessary to use the fins on platforms or offer a dimensionless metric that can be used to compare the efficiency of different designs between different UUV control systems. We generate a dimensionless figure of merit (FOM) that can be used for creating a comparative to other designs and control systems and optimizing power efficiency. 

To generate a FOM, we develop forward models with mathematical and machine learning approaches that use static and dynamic information about the fin design and kinematics on the control system to output a low-power, time-efficient, and accurate prediction of power consumption. Combined with another integrated model for determining thrust \cite{lee_data_2022}, the model can take in a desired thrust or location to find the most power-efficient fin kinematics to accomplish its goal. 

We use the validated thrust and power model to output kinematics interpolations across the entire possible gait space, allowing us to analyze FOM trends and develop a forward FOM model that can predict the efficiency of a certain gait. We generate a FOM that accomplishes the following objectives:
\begin{enumerate}
    \item Adaptability: the FOM forward model can function even if physical properties (actuator, fin design) are switched by simply retraining the model
    \item Practicality: the FOM can accurately model and account for an actuator's power loss
    \item Flexibility: the FOM can optimize for different outputs or forces by simply feeding in a different model
\end{enumerate}

\section{Materials and Methods}
The research objectives guiding this study are to identify the most important parameters that affect flapping fin power consumption and efficiency to develop a forward model of propulsive efficiency for practical comparison and analysis.

\subsection{Parameters and Outputs}
There is one fin design parameter in this study, material flexibility, and there are three fin materials in total (Table \ref{tab:mat}). There are also four kinematics parameters that define a singular gait, including stroke phase offset, stroke amplitude, pitch amplitude, and flapping frequency (Table \ref{tab:param}). 

We collect experimental data on forces, current draw and voltage for the different fin gaits, which we use to develop our forward models for force and power. This information is laid out in Table \ref{tab:out}, and force vectors are defined in Figure 1. 

\begin{table}[ht]
  \caption{Fin Material Properties}
  \label{tab:mat}
  \centering 
  \begin{threeparttable}
    \begin{tabular}{p{0.4\linewidth} p{0.4\linewidth}}
     \midrule
     \midrule
    \textbf{Material} & \textbf{Young's Modulus}  \\
     \midrule
     \midrule
    \cmidrule(l r ){1-2}
    Rigid Nylon & ~1 GPa \\
    \cmidrule(l r ){1-2}
    PDMS 1:10 & 850 kPa \\
    \cmidrule(l r ){1-2}
    PDMS 1:20 & 310 kPa \\
    \midrule\midrule
    \end{tabular}
\end{threeparttable}
  \end{table}
  
\begin{table}[ht]
  \caption{Kinematic Gait Parameters}
  \label{tab:param}
  \centering 
  \begin{threeparttable}
    \begin{tabular}{>{\raggedright\arraybackslash}p{0.275\linewidth} p{0.1\linewidth} p{0.5\linewidth}}
     \midrule
     \midrule
    \textbf{Parameter} & \textbf{Symbol} & \textbf{Description} \\
     \midrule
     \midrule
    \multicolumn{3}{c}{\textit{Static Kinematics}} \\
    \cmidrule(l r ){1-3}
    Frequency (Hz) & $f$ & Number of flap cycles per second \\
    \cmidrule(l r ){1-3}
    Stroke Pitch Offset ($\degree$) & $\delta$ & Phase offset of the pitch cycle to the stroke cycle, calculated as $\frac{1}{16}$th of one cycle \\
    \cmidrule(l r ){1-3}
    Stroke Amplitude ($\degree$) & $\Phi$ & Maximum stroke angle over one flap cycle \\
    \cmidrule(l r ){1-3}
    Pitch Amplitude ($\degree$) & $\Theta$ & Maximum pitch angle over one flap cycle \\
    \cmidrule(l r ){1-3}
    \multicolumn{3}{c}{\textit{Dynamic Kinematics}} \\
    \cmidrule(l r ){1-3}
    Stroke Angle ($\degree$) & $\phi$ & Time history of stroke angle \\
    \cmidrule(l r ){1-3}
    Pitch Angle ($\degree$) & $\theta$ & Time history of pitch angle \\
    \midrule\midrule
    \end{tabular}
\end{threeparttable}
  \end{table}
 
\begin{figure}[h!]
\centerline{\includegraphics[width=0.95\linewidth,height=\textheight,keepaspectratio]{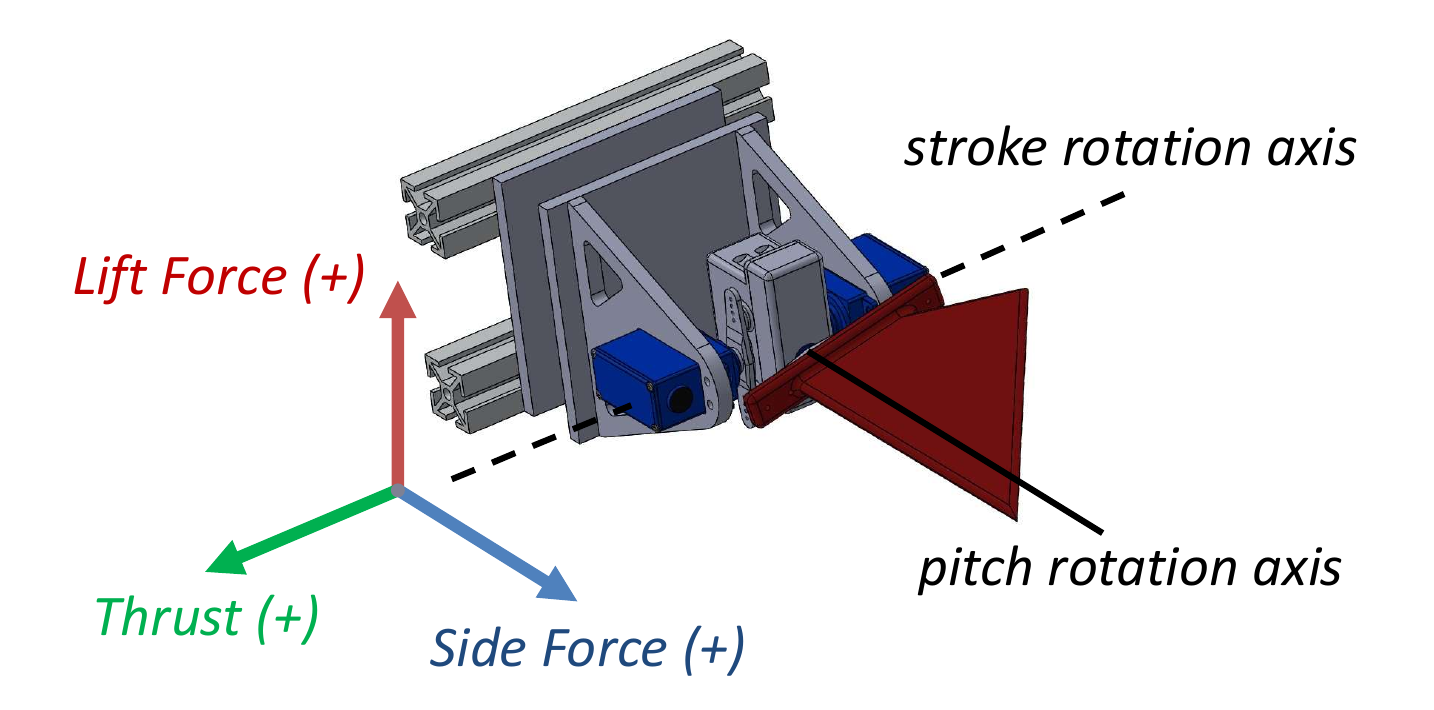}}
\caption{CAD design of single flapping fin propulsor with coordinate reference frames.}
\label{fig}
\end{figure}
 
\begin{table}[h!]
  \caption{Control System Measurements}
  \label{tab:out}
  \centering 
  \begin{threeparttable}
    \begin{tabular}{>{\raggedright\arraybackslash}p{0.22\linewidth} p{0.1\linewidth} p{0.555\linewidth}}
     \midrule
     \midrule
    \textbf{Parameter} & \textbf{Symbol} & \textbf{Description} \\
     \midrule
     \midrule
    \multicolumn{3}{c}{\textit{XYZ Forces}} \\
    \cmidrule(l r ){1-3}
    Thrust (N) & $T$ & Force generated along stroke axis \\
    \cmidrule(l r ){1-3}
    Lift (N) & $L$ & Force generated perpendicular to both \\
    \cmidrule(l r ){1-3}
    Side Force (N) & $S$ & Force generated along pitch axis \\
    \cmidrule(l r ){1-3}
    \multicolumn{3}{c}{\textit{Power Consumption}} \\
    \cmidrule(l r ){1-3}
    Stroke Current (A) & $I_{\phi}$ & Time history of current draw for the stroke actuator \\
    \cmidrule(l r ){1-3}
    Pitch Current (A) & $I_{\theta}$ & Time history of current draw for the pitch actuator \\
    \cmidrule(l r ){1-3}
    Voltage ($V$) & $V$ & Voltage of both actuators \\
    \midrule\midrule
    \end{tabular}
\end{threeparttable}
  \end{table}

\subsection{Data Collection}

The parameter space for the kinematics variables is large, but we constrain the tests to parameter values that are physically achievable, given an operating frequency. At higher frequencies, the fins are physically unable to reach certain stroke and pitch amplitudes. Equation \ref{eq:1} defines our range of achievable strokes and pitches with respect to frequency: 
\begin{equation}
    \label{eq:1}
    \text{Attainable gaits:} \begin{cases}
        \text{$0<\Phi<97-f*30$}\\
        \text{$0<\Theta<75-f*26$}\\
    \end{cases}  
\end{equation}\\
With this equation, we collected data that covers the entire scope of the achievable kinematics range at each frequency. In total, 864 unique gaits listed in Table \ref{tab:combo} were tested for each design. For each fin gait, 10 flap cycles were run, and 5 of the middle cycles were used for analysis to account for discrepancies when the actuator started and ended the cycle motions. Data was collected in a zero velocity flow condition, which previous research has demonstrated is important for low-speed maneuvering to station-keep and offset buoyancy \cite{geder_fluid-structure_2021}.

\begin{table}[ht]
  \caption{Gait Combinations}
  \label{tab:combo}
  \centering 
  \begin{threeparttable}
    \begin{tabular}{>{\raggedright\arraybackslash}p{0.325\linewidth} p{0.575\linewidth}}
     \midrule
     \midrule
    \textbf{Parameter} & \textbf{Symbol}  \\
     \midrule
     \midrule
    Stroke Amplitude ($\degree$) & $0, 15, 25, 32.5, 40, 55$  \\
    \cmidrule(l r ){1-2}
    Pitch Amplitude ($\degree$) & $0, 15, 25, 32, 38, 55$ \\
    \cmidrule(l r ){1-2}
    Frequency (Hz) & $0.75, 1.00, 1.25, 1.50, 1.75, 2.00$ \\
    \cmidrule(l r ){1-2}
    Stroke Pitch Offset ($\degree$) & $-22.5, 0, 22.5, 45$  \\
    \cmidrule(l r ){1-2}
    Voltage ($V$) & Constant at $4.98V$ \\
    \midrule\midrule
    \end{tabular}
\end{threeparttable}
  \end{table}
  
This process was replicated for all three fin material designs. Taking the experimental data as defined in Table \ref{tab:out}, we compute the total power consumption of both the stroke and pitch actuators with Equation \ref{eq:2}.
\begin{equation}
    \label{eq:2}
    P = I_{\phi} * V + I_{\theta} * V
\end{equation}

\subsection{Experimental Setup}
\begin{figure}[h]
\centerline{\includegraphics[width=\linewidth,height=\textheight,keepaspectratio]{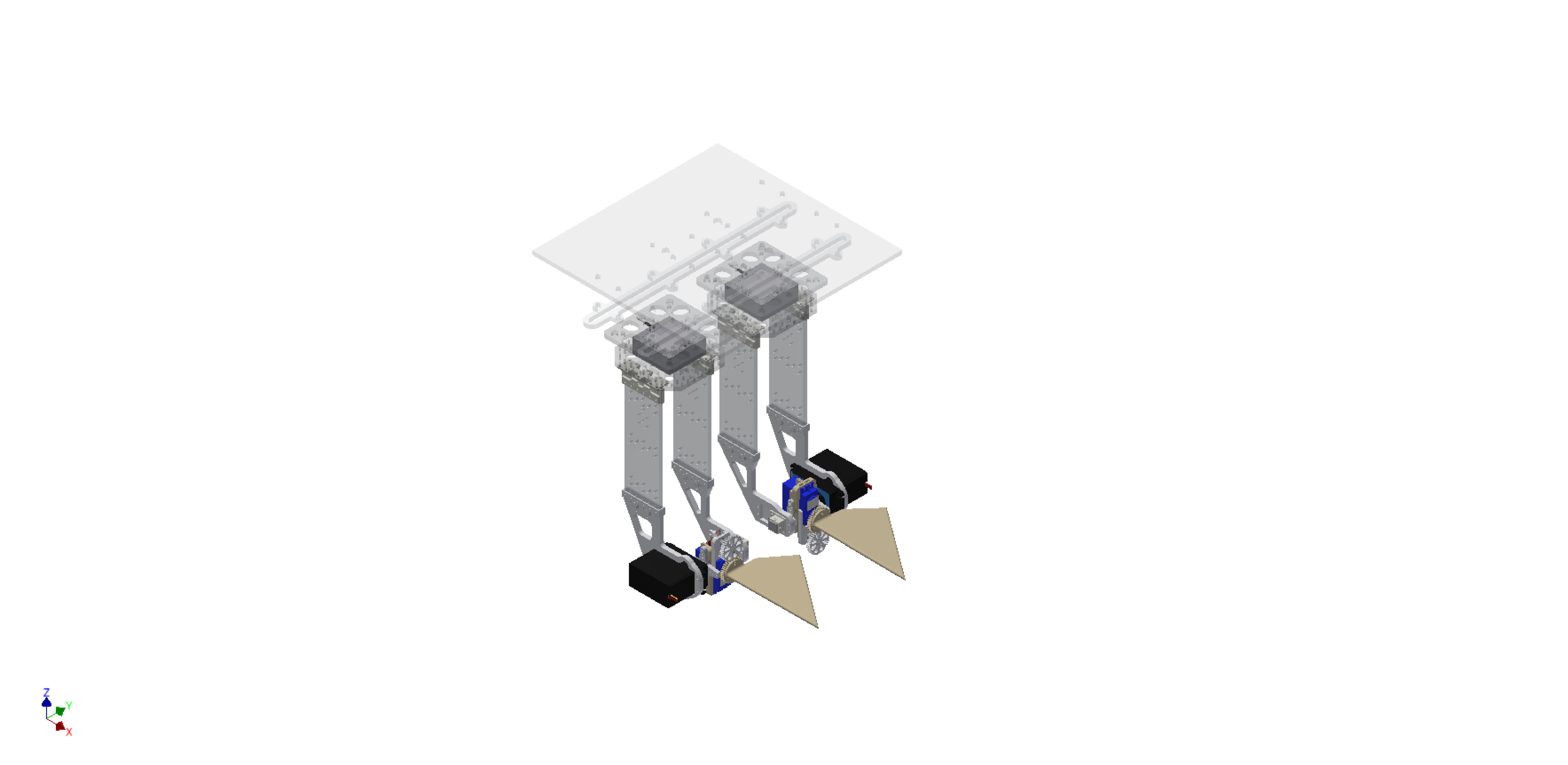}}
\caption{CAD design of tandem fins mounted to instrumentation and control platform.}
\label{fig:cad-multifin}
\end{figure}

\begin{figure}[h]
\centerline{\includegraphics[width=\linewidth,height=\textheight,keepaspectratio]{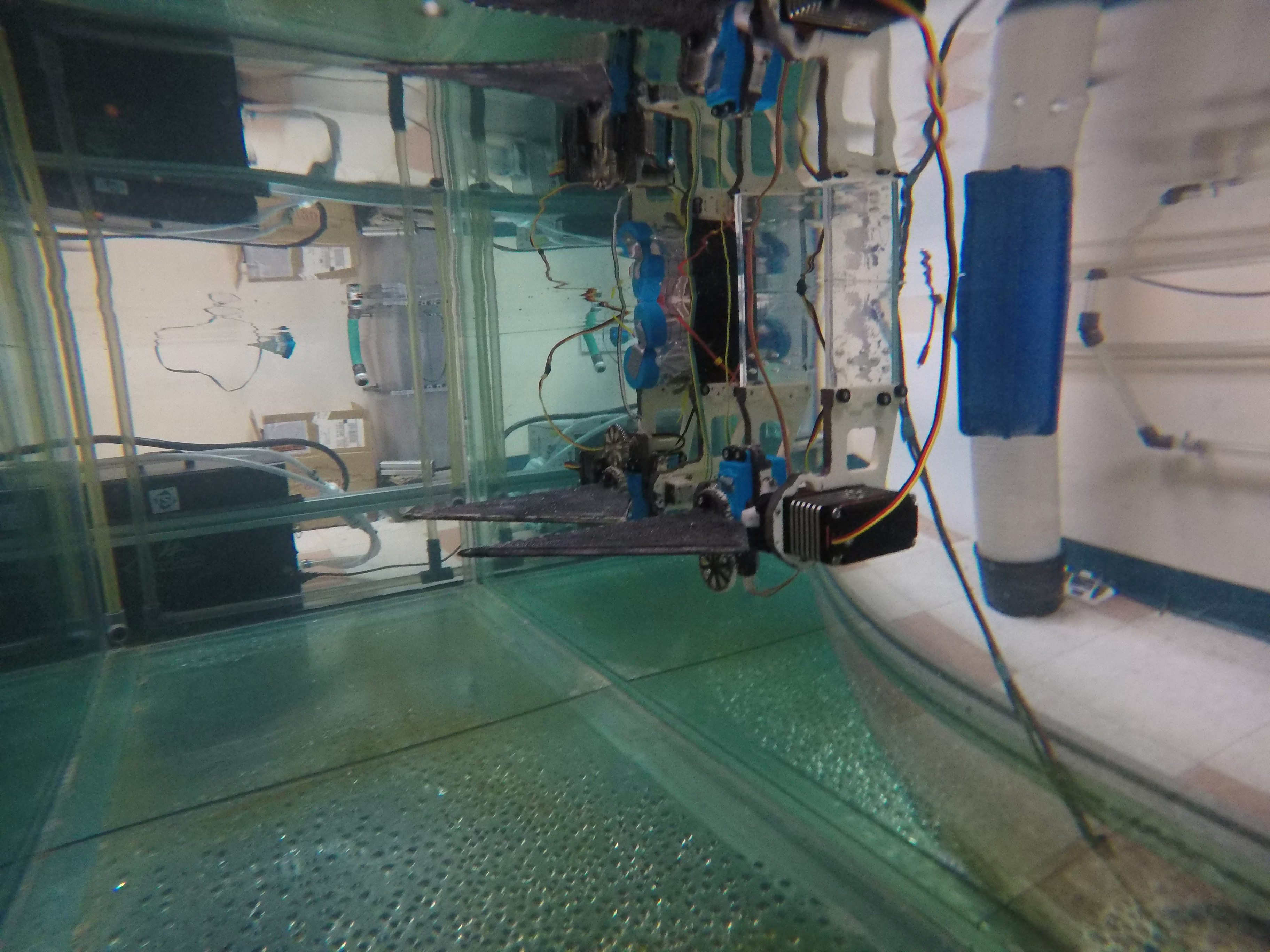}}
\caption{Tandem fin platform in experimental test environment.}
\label{fig:tank}
\end{figure}

Data was collected on the experimental setup shown in Figures \ref{fig:cad-multifin} and \ref{fig:tank}. As prior research has demonstrated that tandem fin configurations have minimal effects on thrust output and power consumption,
% [INSERT HERE],
we only run single-fin tests to gather data. 

%Tests are conducted in a zero free-stream flow condition which is an important operating point for hover-capable UUVs as they station-keep in shallow waters and offset buoyancy at low speed [4], [27], [28].

%Experiments are conducted in a 2.41 x 0.76 x 0.76 m (length x width x height) transparent glass tank.  Each fin is mounted to a platform connected to a pair of shared rails through locking slider bars so they can be set a defined distance apart.  The fin actuators are controlled through a microcontroller that is programmed to allow different fin motions to be tested.  Potentiometers (TT Electronics P260) are used to measure the angular motion of the stroke and leading-edge pitch of each fin, and three-axis load cells (Interface 3A60A) are used to measure forces generated by each fin. 

The control platform was mounted in a 2.41 x 0.76 x 0.76m (length, width, height) glass tank. A microcontroller controlled the fin actuators to collect data on the programmed gait combinations as laid out in Tables \ref{tab:param} and \ref{tab:combo}. Potentiometers (TT Electronics P260) measure the stroke and pitch angles over time, while load cells (Interface 3A60A) measure the generated forces \cite{geder_fluid-structure_2021}.

\section{Model}
\subsection{Objectives}
Developing a forward model that inputs gait information and outputs a thrust or power prediction allows for a wider range that we can interpolate results from our figure of merit. Additionally, a fast enough model can allow for future integration onto the control system to take in power or the FOM as a metric in an inverse model. These objectives guided the selection and development of our model:
\begin{enumerate}
    \item Completion of a baseline model that can accurately take in kinematics data (frequency, stroke amplitude, pitch amplitude, and offset) to output predicted power
    \item Capability to take in static information such as material and flexibility to use different models to maximize accuracy and usefulness of integrated model
    \item Run-time speed of 100 forward passes per second at minimum
\end{enumerate}

First, the model takes in design-related parameters such as material, design, rigidity, and tandem fin spacing to procure the most optimal model. Since the fins installed are static and will not be replaced during missions, this allows the model to only load what is relevant to the mission without wasting excess computational power. Second, the loaded model will take in more dynamic and changing kinematics-related information such as frequency, stroke angle, pitch angle, angle offset, and tandem phasing.

\subsection{Model Results}
We examined five approaches to model power; two were polynomial models and three utilized ML approaches. 

\begin{figure}[h!]
\centerline{\includegraphics[width=\linewidth,height=\textheight,keepaspectratio]{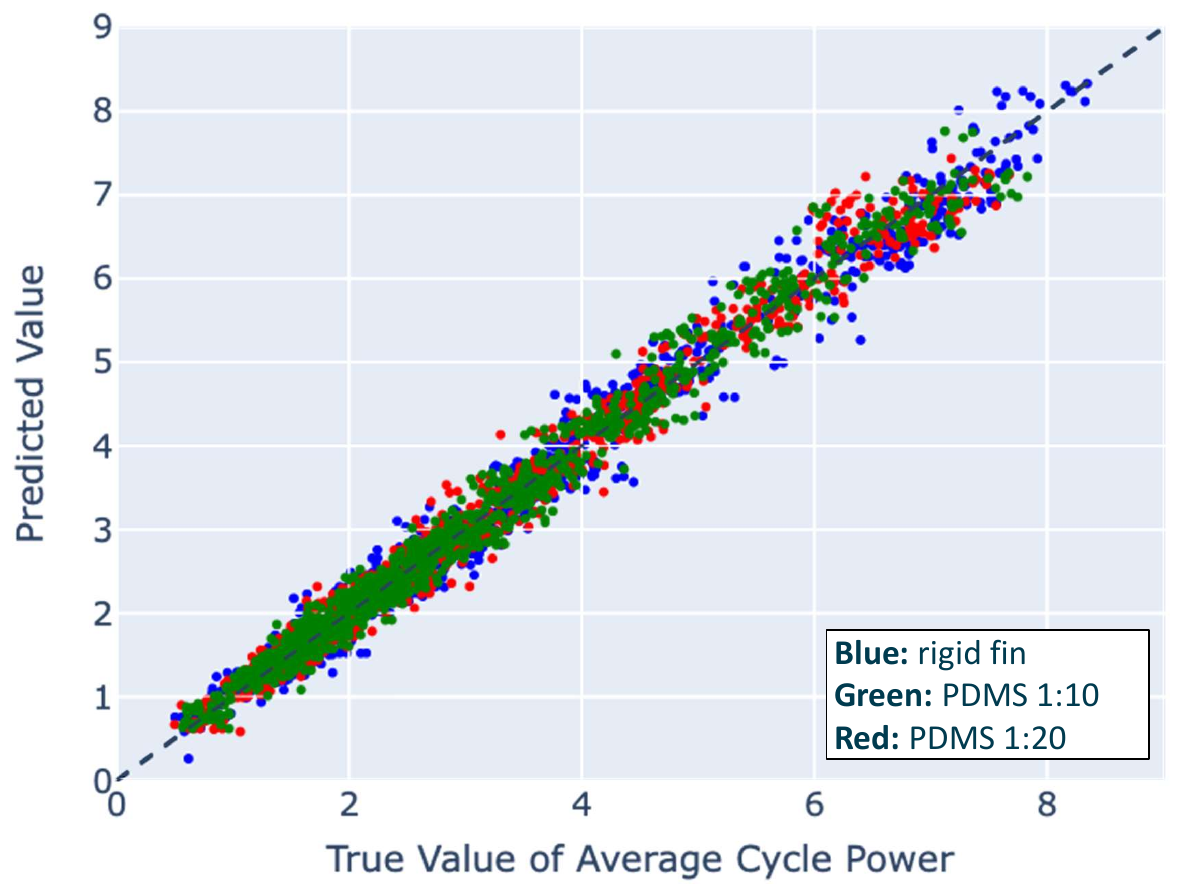}}
\caption{Quartic polynomial model performance for synthetic data, comparing predicted power values to the true power value. Colors indicate the material, with blue (rigid) green (1:10) red (1:20)}
\label{fig:model}
\end{figure}

Our baseline model, the linear model, performed the fastest with an average error across all 3 data sets of 0.3891 W.
% [INSERT HERE deviation of the three data sets?]
As the true value and predicted values appeared to have an exponential relationship in the linear model, we tested a quartic polynomial which produced better results (Figure \ref{fig:model}) with an average error of 0.1815 W. Out of the three data sets, the PDMS 1:10 fins fit the best to the quartic regression.

Utilizing a Convolutional Neural Network (CNN) as a baseline ML approach, we see an increase in accuracy with an average error of only 0.0907 W while still being able to operate at a speed fast enough for model integration. 

While the first three models were implemented and other approaches such as a Multi Layer Perceptron Regression were considered, we ultimately chose to use a time-series model for modeling thrust and power to understand patterns present throughout a gait. As such, we used a Long Short-Term Memory model (LSTM) similar to the thrust LSTM from previous literature \cite{lee_data_2022} that is able to substantially decrease the average error between interpolated data and create a time series of power consumption during one flap cycle. Most importantly for our purposes, the LSTM can accurately interpolate between gaps of data, which is useful for our FOM analysis.

When trained on all experimental gaits, the thrust LSTM reached an average error of 0.0083 N and the power LSTM reached an average error of 0.0408 W. The most visible shortfall of the power LSTM model is its inability to grasp the time history of power consumption at certain gaits. At both extremes of gaits with lower amplitudes or higher amplitudes, the power history of each cycle can be vastly different, leaving the model unable to fully understand the time history. For the purpose of interpolating the power consumption across a gait, this is not a setback.

Out of all five power models, the LSTM has the best performance but is also the most time-consuming. Table \ref{tab:model} highlights the results and relative time performances for a forward model.

\begin{table}[ht]
  \caption{Model Performances}
  \label{tab:model}
  \centering 
  \begin{threeparttable}
    \begin{tabular}{>{\raggedright\arraybackslash}p{0.3\linewidth} p{0.3\linewidth} p{0.275\linewidth}}
     \midrule
     \midrule
    \textbf{Model} & \textbf{Averaged Error (W)} & \textbf{Runtime}  \\
     \midrule
     \midrule
    Linear Polynomial & 0.3891 & Low  \\
    \cmidrule(l r ){1-3}
    Quartic Polynomial & 0.1815 & Low \\
    \cmidrule(l r ){1-3}
    MLP & 0.1229 & Medium \\
    \cmidrule(l r ){1-3}
    CNN & 0.0907 & Medium \\
    \cmidrule(l r ){1-3}
    LSTM & 0.0072 & High \\
    \midrule
    \midrule
    \end{tabular}
\end{threeparttable}
  \end{table}

\subsection{Generating Models}
%BEFORE PUBLISH : Should this be generating interpolations still? if so, needs additional content
With the LSTM having the most accurate interpolations, we train each LSTM model to 1000 epochs for modeling both power and thrust predictions. For the thrust model, we utilize a previously developed model on the rigid fin data set \cite{lee_data_2022} and tune it to produce optimal results on the PDMS 1:10 and 1:20 data sets. The statistics for the LSTM models produced are found in Table \ref{tab:LSTM}.

\begin{table}[ht]
  \caption{Average Error for LSTM Models}
  \label{tab:LSTM}
  \centering 
  \begin{threeparttable}
    \begin{tabular}{p{0.3\linewidth} p{0.15\linewidth} p{0.15\linewidth} p{0.15\linewidth}}
    %{m{15mm} m{70mm} m{18mm}}
     \midrule
     \midrule
    \textbf{Interpolations} & \textbf{Rigid} & \textbf{PDMS 1:10} & \textbf{PDMS 1:20}  \\
     \midrule
     \midrule
    Power (W) & 0.0236 & 0.072 & 0.0268 \\
    \cmidrule(l r ){1-4}
    Thrust (N) & 0.0002 & 0.0186 & 0.0041 \\
    \midrule
    \midrule
    \end{tabular}
\end{threeparttable}
  \end{table}

\section{Figure of Merit}
\subsection{Objectives}
We create a dimensionless FOM that takes in a gait or flap cycle along with relevant design information and outputs a metric that rates the efficiency of force generation.

\subsection{Development}
A baseline FOM simply computing the force in question over the power would provide a relative metric that can be used to compare gaits assuming the same flow speed, displayed in Equation \ref{eq:fom1}. For integration purposes, this would be suitable and could return as the pure value or a percentage compared to the highest recorded gait efficiency for a loaded design. 

\begin{equation}
    \label{eq:fom1}
    \eta  = \frac{F_{avg}}{P_{avg}}
\end{equation}\\

While this FOM has units, we can convert it to a dimensionless parameter by multiplying by a velocity. Since all tests were conducted in a constant, static flow, this FOM velocity term can be set to 1 m/s, as shown in Equation \ref{eq:fom2}, effectively cancelling the velocity term for relative efficiency comparisons between our tests.
\begin{equation}
    \label{eq:fom2}
    \eta  = \frac{F_{avg}*1\text{ m/s}}{P_{avg}}
\end{equation}\\

To make our FOM dimensionless for all future flow conditions, we can iterate off a previous FOM
% [INSERT HERE ravi citation] %BEFORE PUBLISH
to develop a universal FOM that implements the flow speed by calculating the tip velocity, as shown in Equations  \ref{eq:fom3} and \ref{eq:fom4}. 
\begin{equation}
    \label{eq:fom3}
    \eta  = \frac{F_{avg}v}{P_{avg}}
\end{equation}
\begin{equation}
    \label{eq:fom4}
    v  = \frac{F_{avg}}{P_{avg}}
\end{equation}\\

This FOM is easily able to adapt to different objectives. Because we collect data for all force axes and the individual stroke and pitch actuators, we are able to create new deviations of Equation \ref{eq:fom3} to compare an individual fin efficiency or how efficient a gait is for a specific force axis (Equation \ref{eq:forces}). The FOM can also be used to isolate how efficiency changes when changing individual actuators (Equation \ref{eq:angle}), and optimize for different objectives in future integration. Examples are listed below:\\
% \begin{center}
$T$ = Thrust, $L$ = Lift, $S$ = Side Force
\begin{equation}
    \label{eq:forces}
    \eta_{Ft}  = \frac{T_{avg}v}{P_{avg}};\quad
    \eta_{Fl}  = \frac{L_{avg}v}{P_{avg}};\quad
    \eta_{Fs}  = \frac{S_{avg}v}{P_{avg}} 
\end{equation}
$P_{s}$ = Power consumed by stroke actuator\\
$P_{p}$ = Power consumed by pitch actuator
\begin{equation}
    \label{eq:angle}
    \eta_{As}  = \frac{F_{avg}v}{P_{s}};\quad
    \eta_{Ap}  = \frac{F_{avg}v}{P_{p}}
\end{equation}\\

% \end{center}

%results: describe the models themselves in their architecture section but pros and cons and why x>y in results
%appendix? if it doesn't relate to the core, if model comparison is part of it put it in methods/model and show a comparative in results. if it's not, bump it into an appendix or documentation that doesn't go in the paper

\section{Results}
\begin{figure}[htbp]
\label{fig:rigid_interp}
\centerline{\includegraphics[width=\linewidth,height=\textheight,keepaspectratio]{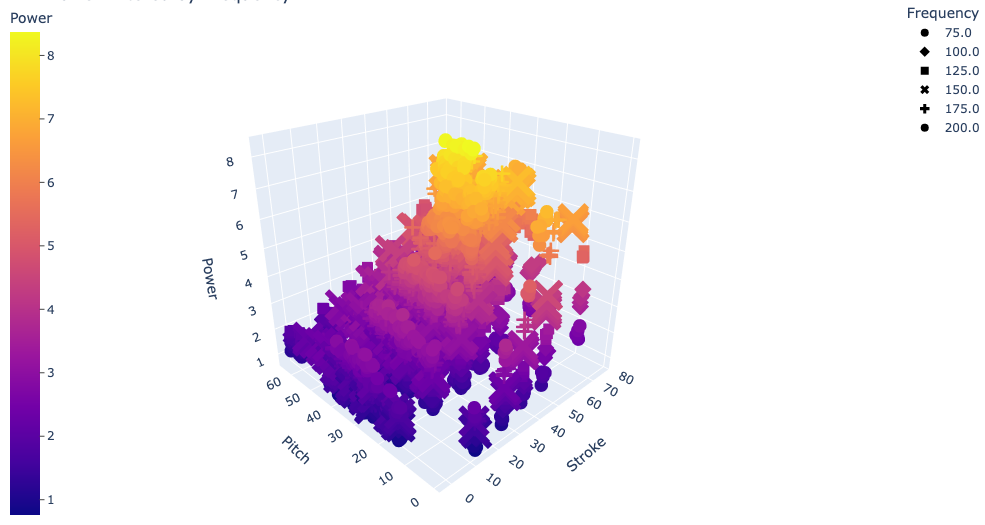}}
\caption{3D visualization of the rigid fin data set.}
\label{fig:interps}
\end{figure}

We used our generated LSTM models to generate predictions at points interpolated between training data. In total, we generated interpolations within the constraints given by our collected data. We interpolated data for every stroke and pitch combination from 0 to 55 degrees with 1 degree increments, frequency from 0.75Hz to 2Hz with 0.125Hz increments, and SPO from -22.5 to 45 degrees with 5.625 degree increments. In total, 435,600 interpolations were calculated for each data set. A sample of the rigid fin data space is shown in Figure \ref{fig:interps}. The interpolations filled gaps of data.

\begin{figure*}[p]
\centering
\includegraphics[width=\linewidth]{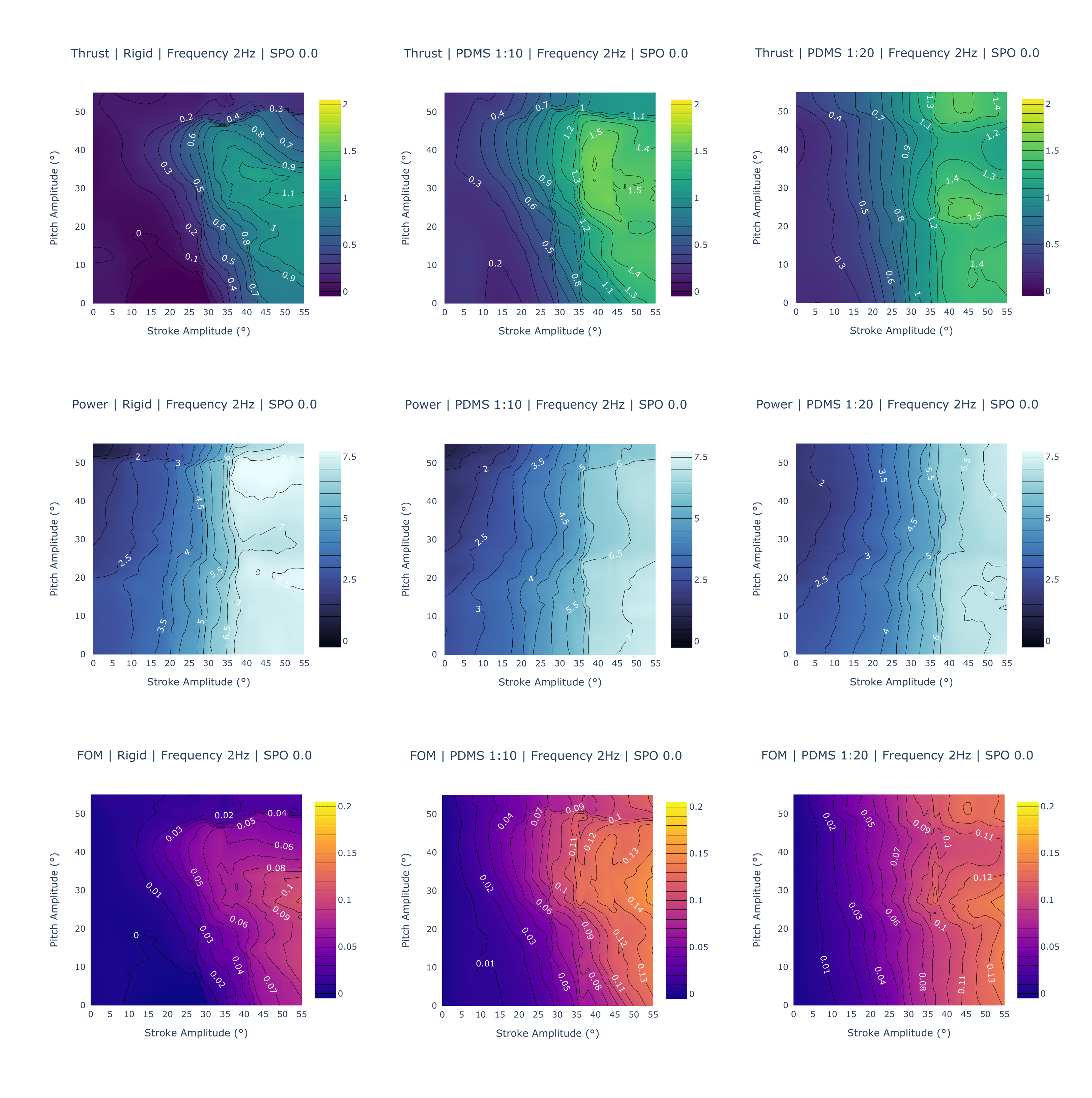}
\caption{Figure of merit, thrust, and power contours. The columns are different data sets (rigid, PDMS 1:10, and PDMS 1:20) while the rows graph different gait results (FOM value, Thrust, and Power). The PDMS 1:10 fin generates the highest possible thrust and is higher overall in more gaits; the rigid fin is significantly worse at thrust generation than either of the PDMS fins. The PDMS 1:10 and 1:20 fins are comparable in power consumption but differ in trends at higher stroke and pitch combinations; the rigid fin consumes significantly more power. The PDMS 1:10 fin has the largest FOM values, with the PDMS 1:20 fin following. The rigid fin is significantly worse in all 3 metrics.}
\label{fig:fom}
\end{figure*}

\begin{figure*}[t!]
\centering
\includegraphics[width=\textwidth,height=\textheight,keepaspectratio]{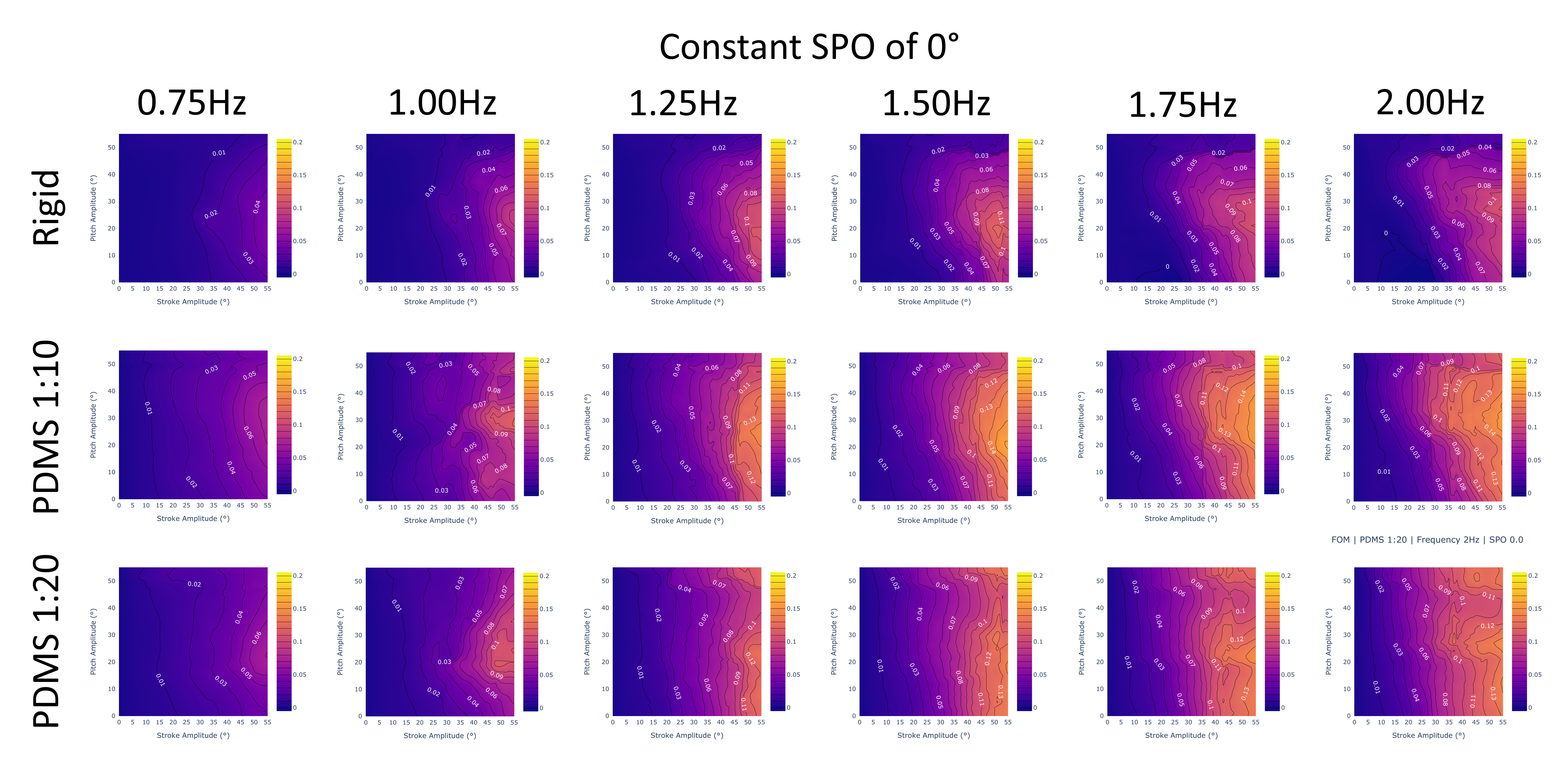}
\caption{Figure of Merit results for a constant Stroke-Pitch-Offset of 0 degrees and varying frequency. The rows graph all interpolated frequency values with 0.25Hz increments while the columns are different data sets (rigid, PDMS 1:10, and PDMS 1:20). We verify that PDMS 1:10 is the best performing across all frequencies with higher FOM values and averages compared to the PDMS 1:20 or rigid fins. Additionally, increasing the frequency will improve the FOM regardless of fin design.}
\label{fig:freq}
\end{figure*}

To gain a better understanding of how our Figure of Merit correlates to the thrust force and power consumption, we created a grid of contours for one gait, shown in Figure \ref{fig:fom}. Here, the stroke and pitch range for a frequency of 2 Hz and Stroke-Pitch Offset of 0$^\circ$ is depicted. The grid contains columns with the three material data sets and each row graphs part of the figure of merit equation: FOM, thrust, and power in that order. 

Beginning with the thrust, a few trends are immediately apparent. First, the fin design that can generate the largest force is the PDMS 1:10 design, generating a maximum force of around 1.6 N at a 40$^\circ$ stroke amplitude and 30-40$^\circ$ pitch amplitude. Both other designs trail behind, with the PDMS 1:20 fin being able to generate a thrust of 1.5 N and the rigid fin only managing up to 1.1 N. The PDMS 1:10 fin also has more gaits at higher levels. An observation of the contour reveals that there are more combinations of stroke and pitch that produce higher thrusts when compared to the PDMS 1:20 fin. An analysis of all gaits including other SPO and frequency confirms these findings. The PDMS 1:10 fin's maximum thrust is 2.1 N, while the PDMS 1:20 fin produces a maximum force of 1.6 N and the rigid fin produces a maximum force of 1.2 N. The PDMS 1:10 fin has the highest average thrust generation, followed by the PDMS 1:20 fin.

Power consumption goes in the reverse order. The design that requires the highest wattage is the rigid fin design, requiring 7.6 W for any gait with a stroke amplitude of more than 40$^\circ$. The PDMS 1:10 and PDMS 1:20 fins are similar, with a maximum power consumption of 7.1 W in this contour. However, at cases above 40$^\circ$ stroke and 30$^\circ$ pitch, the PDMS 1:10 fin observes lower wattage consumed at the same gait combination. The PDMS 1:20 fin appears to depend less on the pitch amplitude, with power more dependent on the stroke amplitude. These observations are verified when looking at all gaits. While the PDMS fins have a maximum wattage of around 7.5 W, it occurs at much fewer gaits than with the rigid data set. 

Another interesting trend is visible when comparing the FOM and thrust charts, which appear almost identical with only a few differences in their trends. The explanation becomes evident when looking at the contours for power, which have a near-linear trend across stroke amplitude. While pitch amplitude does affect both the PDMS 1:10 and 1:20 designs, the stroke amplitude has the most recognizable and significant effect. Future work will include generating additional figures to verify that this trend exists across all frequencies and stroke-pitch offsets.

\begin{figure*}[t!]
\centering
\includegraphics[width=\textwidth,height=\textheight,keepaspectratio]{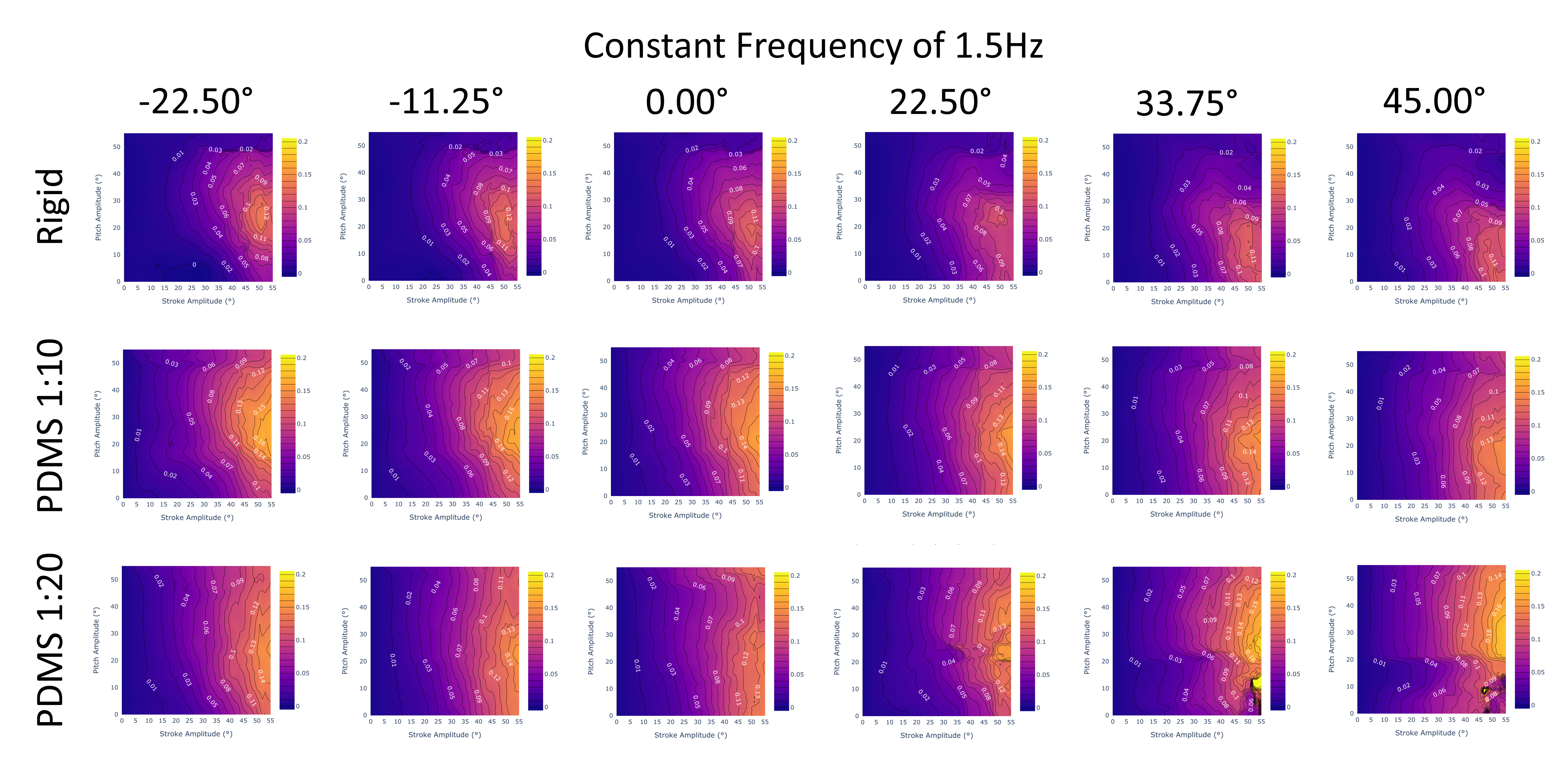}
\caption{Figure of Merit results for a constant frequency of 1.5Hz and varying Stroke-Pitch-Offset. The rows graph all interpolated SPO values with 11.25$^\circ$ increments while the columns are different data sets (rigid, PDMS 1:10, and PDMS 1:20). We verify that the PDMS 1:10 fin is the best performing across all SPO values with higher FOM values and averages compared to the PDMS 1:20 or rigid fins. With the exception of 22.5-45 degrees for PDMS 1:20, a more negative SPO will slightly improve the FOM regardless of fin design.}
\label{fig:spo}
\end{figure*}

This analysis allows us to conclude that the PDMS 1:10 fin design is the most efficient out of all 3 designs, with the highest thrust generation and efficiency. Following in second is the PDMS 1:20 fin, which has the second largest thrust generation and efficiency. These trends are confirmed across the ranges of stroke and pitch (Figure \ref{fig:fom}) as well as frequency (Figure \ref{fig:freq}) and SPO (Figure \ref{fig:spo}). This suggests that the most efficient design that is able to generate the largest thrust may lie between the two, and is something of interest for future exploration.  

From Figure \ref{fig:freq}, we observe that across the entire range of frequencies, the PDMS 1:10 outperforms both the rigid and PDMS 1:20 fin designs. Additionally, we observe that regardless of fin design, increasing the frequency will improve the FOM metric.

From Figure \ref{fig:spo}, we observe that across the entire range of stroke-pitch offset, the PDMS 1:10 outperforms both the rigid and PDMS 1:20 fin designs. Additionally, we observe that regardless of fin design, a more negative offset will slightly improve the FOM metric, although the difference is very marginal. At high SPO, the PDMS 1:20 fin design appears to diverge from the expected trends at 22.5 $^\circ$ or higher and invites future exploration. In the future, revisiting the model's training data for high SPO will likely resolve the issue. 

We conclude the following from the data interpolated: 
\begin{enumerate}
    \item The best performing fin design is PDMS 1:10, followed by PDMS 1:20. Both consume similar amounts of power, but PDMS 1:10 fins produce a higher thrust. The rigid fin consumes more power and produces less thrust.
    \item The most optimal fin design likely lies in between the PDMS 1:10 and 1:20 fins.
    \item The most efficient gait will occur at a high stroke (40-55), centered pitch (20-35), high frequency (2Hz) and low SPO (-22.5$^\circ$). 
\end{enumerate}

% [INSERT HERE ] 

\section{Conclusion and Future Work}
We use a forward-passing LSTM model to generate kinematic interpolations and for fin gaits with the goal of integration onto a control system to optimize for the efficiency of gaits and provide a better understanding of material designs and their relation to efficiency. 

We evaluate four different models for thrust and power to interpolate between experimental data with high accuracy: the linear model, quartic polynomial model, Convolutional Neural Network, and Long-Short-Term Memory model. All four evaluated models accomplish all three criterion we laid out. They are able to:
\begin{itemize}
    \item Complete a baseline model that inputs gait parameters to output either thrust or power with minimal error
    \item Retrain on different designs and maintain a similar or better accuracy
    \item Run at a speed and size suitable for integration onto a control system ($>$100 computations per second)
\end{itemize}
Out of all four models, we find that the LSTM model is able to produce the most accurate results on both the full data set and when we remove specific gaits to create interpolations. 

Using the generated interpolations, we develop a dimensionless Figure of Merit that is able to compare our fin efficiency to other flapping systems and evaluate the efficiency of gaits onboard our control system. 

With the FOM, we conclude that both PDMS materials are more efficient than the rigid fin, with the PDMS 1:10 fin generating the maximum thrust with the lowest power consumption. There were observable trends consistent across all materials, with a higher frequency, stroke amplitude, and negative Stroke-Pitch-Offset all contributing to a greater efficiency. The most efficient gait was concluded to be for the PDMS 1:10 data set with a -22.5$^\circ$ Stroke-Pitch-Offset and frequency of 2 Hz. This understanding will allow us to both design fins that generate a higher thrust and maintain the highest power efficiency, and tune an inverse search model to search gaits it knows to be power-efficient or optimal. 

In the future, we aim to integrate the inverse search model with the FOM as a weighting onto the control system to generate the most efficient or powerful gaits. Before this integration is complete, we will combine both the thrust and power LSTM models to decrease the model's compute time on the control system. Additionally, we will further verify our FOM by introducing physical bounds on the thrust/power values a model can predict and conducting additional tests with non-zero flow speeds. Introducing disturbances and turbulence into the water ahead of time will aid in evaluation of model efficiency prediction accuracy for dynamic environments similar to those the UUV will swim in. 

% The data gathered from these trials primarily come from two sources: high-fidelity computation fluid dynamics (CFD) simulations, and experimental rigs in highly controlled and monitored environments. Problematically, both methods avoid including turbulence and noise to ensure the accuracy of testing results and simulations to design the fin: experimental rigs measure data in carefully monitored environments and CFD simulations aim to only model a design's performance in ideal environments. As such, both do not accurately represent the unpredictable nature of a UUV's near-surface domain. 

%  \section*{Acknowledgment}
% I owe a large thank you to all of the amazing people who have made my research and internship possible. Thank you to my mentor Jason for bringing me on, and Lee and Marius for teaching and guiding me throughout my project, introducing me to concepts and the experimentation. Additionally, thank you Kamal and Julian for chiming in and sharing your thoughts and ideas as my project was taking shape. A final thank you to Akshat and the other interns in the lab, meeting everyone and working at NRL has been one of the most memorable periods of my life. 

%\section*{References}
% \bibliographystyle{aaai22}
\bibliography{references}

%  \section{Acknowledgment}
% I owe a large thank you to all of the amazing people who have made my research and internship possible. Thank you to my mentor Jason for bringing me on, and Lee and Marius for teaching and guiding me throughout my project, introducing me to concepts and the experimentation. Additionally, thank you Kamal and Julian for chiming in and sharing your thoughts and ideas as my project was taking shape. A final thank you to Akshat and the other interns in the lab, meeting everyone and working at NRL has been one of the most memorable periods of my life. 

\end{document}